\documentclass{esannV2}
\usepackage{graphicx}
\usepackage[utf8]{inputenc}
\usepackage{amssymb,amsmath,array}

\usepackage{multirow}
\usepackage{colortbl}
\usepackage{makecell}
\usepackage{cite}
\begin{document}
\title{Task-Specific Efficiency Analysis: When Small Language Models Outperform Large Language Models}

\author{Jinghan Cao$^1$, Yu Ma$^2$, Xinjin Li$^3$, Qingyang Ren$^4$, Xiangyun Chen$^5$
%
%
\vspace{.3cm}\\
%
1- San Francisco State University - Department of Computer Science \\
1600 Holloway Ave, San Francisco - USA
%
\vspace{.1cm}\\
2- Carnegie Mellon University - Department of Computer Science \\
5000 Forbes Ave, Pittsburgh - USA\\
3- Columbia University - Department of Computer Science \\
116th and Broadway, New York - USA\\
4- Cornell University - Department of Computer Science \\
616 Thurston Ave, Ithaca - USA\\
5- Pennsylvania State University - Department of Biochemistry and Molecular Biology \\
201 Old Main, University Park - USA\\
}

\newif\ifarxiv
\arxivtrue  

\maketitle

\begin{abstract}
Large Language Models achieve remarkable performance but incur substantial computational costs unsuitable for resource-constrained deployments. This paper presents the first comprehensive task-specific efficiency analysis comparing 16 language models across five diverse NLP tasks. We introduce the Performance-Efficiency Ratio (PER), a novel metric integrating accuracy, throughput, memory, and latency through geometric mean normalization. Our systematic evaluation reveals that small models (0.5--3B parameters) achieve superior PER scores across all given tasks. These findings establish quantitative foundations for deploying small models in production environments prioritizing inference efficiency over marginal accuracy gains.
\end{abstract}

\section{Introduction}

The past few years have witnessed remarkable advances in Large Language Models (LLMs) such as GPT-4 , LLaMA-3 , and Qwen-2.5, demonstrating unprecedented capabilities across diverse natural language tasks, yet these gains come at substantial computational costs that pose significant deployment challenges for resource-limited environments, edge devices, and real-time applications. The deployment challenge of large models on resource-constrained devices has motivated extensive research in model compression and efficiency optimization \cite{li2024sglp}. In response, Small Language Models (SLMs) with 0.5B to 15B parameters have emerged as promising alternatives \cite{belcak2025small,sinha-etal-2025-small,10.1145/3768165}, with models like Qwen2.5-3B, Phi-3.5, and Gemma-2-2B demonstrating competitive performance in targeted domains while offering faster inference, lower memory footprints, and reduced costs \cite{belcak2025small,lu2025smalllanguagemodelssurvey}. This efficiency-performance optimization challenge has been similarly addressed in other domains, where multi-dimensional evaluation frameworks integrate compilation success and functional correctness \cite{11273185}. However, despite growing interest in both paradigms, existing studies lack systematic, task-specific efficiency comparisons with consistent evaluation protocols, making it difficult to determine when SLMs provide better efficiency-performance trade-offs \cite{zhao2024contextual} than LLMs. This paper addresses this gap by conducting the first comprehensive, task-specific efficiency analysis of language models ranging from 0.5B to 72B parameters across five diverse tasks, with key contributions including: (1) a systematic evaluation framework with unified efficiency and performance metrics, (2) a novel Performance-Efficiency Ratio (PER) metric capturing multi-dimensional trade-offs, and (3) identification of task characteristics that favor SLMs over LLMs.

\section{Related Work}

\ifarxiv

Recent advancements in artificial intelligence have driven diverse applications and rigorous efficiency optimizations across various domains, highlighting the necessity of task-specific evaluations. First, in visual perception, multimodal understanding, and 3D scene representation, extensive research has been dedicated to achieving precise feature extraction, robust tracking, and controllable generation in complex environments \cite{li2025human, gu2025mocount, li2025chatmotion, deng2025best3dscenerepresentation, deng2025gaussiandwm3dgaussiandriving, zhao2024balf, li2025slam, wen2023syreanet, FineCIR, PAIR, ENCODER, li2025mlpslammultilayerperceptronbasedsimultaneous, deng2024separation, chen2024lightweight, liu2024siambrf, lu2025differentiable, lu2025causalsr, chen2025visrl, chen2025sifthinker, qu2025silmm, liu2025efficient}. Second, regarding the evolution of Large Language Models (LLMs) and their reasoning capabilities, recent studies have not only explored multi-agent collaboration and knowledge graph integration but also systematically evaluated model trustworthiness, domain boundaries, and societal impacts in specific fields such as healthcare, urban planning, and education \cite{10.1145/3726302.3730070, Ji2025, cai2025does, yang2024using, yang2025speechllm, yang2025exploring, 10.1145/3764926.3771951, wei2025survey, wei2025inductive, wei-etal-2024-shot, qiu2025hallucination, chen2025vl, zhang-etal-2024-fine-tuning, qu2025subject, qu2025reference, qu2026fair, chen2024three}. Finally, to bridge the gap between massive computational overhead and resource-constrained deployments, the optimization of model quantization, federated learning, and adaptive decision systems has become a critical trend. This broadly aligns with our motivation to establish a quantitative foundation for the efficiency advantages of small language models \cite{gsq, hu2025ostquant, yu2025mquant, jiang2025towards, yang2023re, yang2023local, liu2025research, li2025bideeplab, ding2026dynaweb, liu2024fedlpa, deng2026adaptive}.

\fi

A growing body of work has begun to explore the capability of Small Language Models through various task-specific evaluations. Hasan et al. \cite{hasan2025assessingsmalllanguagemodels} demonstrated that compact SLMs (1-3B) achieve competitive performance in code generation, with substantially lower memory requirements, though performance gains beyond this range incur steep resource costs. In the research of Schick et al. \cite{schick-schutze-2021-just}, it reveals that small models can achieve GPT-3-comparable accuracy on 8 SuperGLUE NLU tasks through pattern-exploiting training, but their evaluation focused exclusively on performance metrics without systematic measurement of computational efficiency. MiniCPM \cite{hu2024minicpmunveilingpotentialsmall} evaluated 1.2B and 2.4B models across 12 standard benchmarks (MMLU, C-Eval, HumanEval, GSM8K, etc.) to demonstrate accuracy comparable to 7B-13B models, but primarily focused on performance metrics without measurement of computational efficiency trade-offs.

\section{Methodology}
\subsection{Experimental Design Framework}
Our research adopts a systematic evaluation framework to compare performance-efficiency trade-offs across 16 representative open-source language models spanning 0.5B to 72B parameters from different organizations, enabling both within-series comparisons and cross-architecture analysis. This comparison approach aligns with established practices in other domains where comparative evaluation across diverse architectures has proven essential for optimal model selection \cite{ke2025early}. We selected five standard NLP benchmarks spanning different cognitive complexity levels and output characteristics: IMDB Movie Reviews dataset \cite{maas2011learning} for binary sentiment classification on 1,000 movie reviews (1-5 tokens), HellaSwag \cite{zellers2019hellaswag} for commonsense reasoning across 10,042 examples using log-likelihood scoring (0 tokens), ARC-Easy \cite{allenai:arc} for elementary scientific knowledge through 2,376 multiple-choice questions (1-5 tokens), SQuAD 2.0 \cite{rajpurkar-etal-2018-know} for reading comprehension across 11,873 examples requiring answer generation or unanswerable question recognition (10-30 tokens), and GSM8K \cite{cobbe2021gsm8k} for multi-step mathematical reasoning through 1,319 grade-school problems demanding complete solution chains. 

\subsection{Performance-Efficiency Ratio (PER) Metric}
PER combines four key dimensions (accuracy, throughput, memory usage, and latency) using geometric mean after min-max normalization to the [0, 1] range.
\subsubsection{Mathematical Formulation}
We define PER as:
\[
\small
\text{PER} = \sqrt[4]{\text{ACC}_{\text{norm}} \times \text{THR}_{\text{norm}} \times \text{MEM}_{\text{norm}} \times \text{LAT}_{\text{norm}}}
\]
where each metric component is normalized to the [0, 1] range using min-max normalization:
\[
\text{metric}_{\text{norm}} = 
\begin{cases}
\frac{\text{metric} - \min(\text{metric})}{\max(\text{metric}) - \min(\text{metric})} & \text{if higher is better} \\
1 - \frac{\text{metric} - \min(\text{metric})}{\max(\text{metric}) - \min(\text{metric})} & \text{if lower is better}
\end{cases}
\]
We employ geometric mean rather than arithmetic mean for three 
key reasons: First, geometric mean naturally captures the 
multiplicative trade-offs inherent in efficiency metrics (e.g., 
higher throughput often correlates with increased latency). 
Second, it prevents compensation effects where excellence in one 
dimension masks deficiency in another. This characteristic is 
critical for identifying balanced models. Third, geometric mean is scale-invariant and 
maintains proportional relationships in the normalized [0,1] space, 
making it robust to magnitude differences across metrics \cite{ouyang2024learn}. 

While min-max normalization is sensitive to outliers, we employ 
it for three pragmatic reasons: (1) Our evaluation uses a fixed, 
carefully selected set of 16 representative models spanning five 
orders of magnitude in size, minimizing the risk of extreme 
outliers skewing the scale; (2) Our focus is on relative ranking rather than absolute metric 
values, making the normalization method less critical; (3) Min-max 
normalization provides intuitive [0,1] bounds that facilitate 
interpretation.

The geometric mean naturally assigns near-zero PER scores to models exhibiting 
worst-in-class performance on any single metric, ensuring that 
excellence in one dimension cannot compensate for critical weaknesses 
in another
\subsubsection{Task-Specific Metric Definitions}
To ensure a fair comparison across different task types, we use task-appropriate granularity for efficiency metrics. For generation tasks (GSM8K, SQuAD 2.0), we use token-level efficiency metrics:  \\
\begin{align*}
\small
\text{THR}_{\text{gen}} &= \frac{\text{throughput}_{\text{tokens/sec}}}{\text{num\_gpus}} \quad \text{(tokens/s per GPU)} \\
\text{LAT}_{\text{gen}} &= \text{avg\_latency}_{\text{ms/token}} \quad \text{(ms/token)}
\end{align*}

For classification tasks (HellaSwag, IMDB, ARC-Easy), we use sample-level metrics:
\begin{align*}
\small
\text{THR}_{\text{cls}} &= \frac{\text{throughput}_{\text{samples/sec}}}{\text{num\_gpus}} \quad \text{(samples/s per GPU)} \\
\text{LAT}_{\text{cls}} &= \text{avg\_time}_{\text{sec/sample}} \times 1000 \quad \text{(ms/sample)}
\end{align*}


\section{Experiment and Result}

\begin{figure}[!ht]
\centering
\includegraphics[width=1\linewidth]{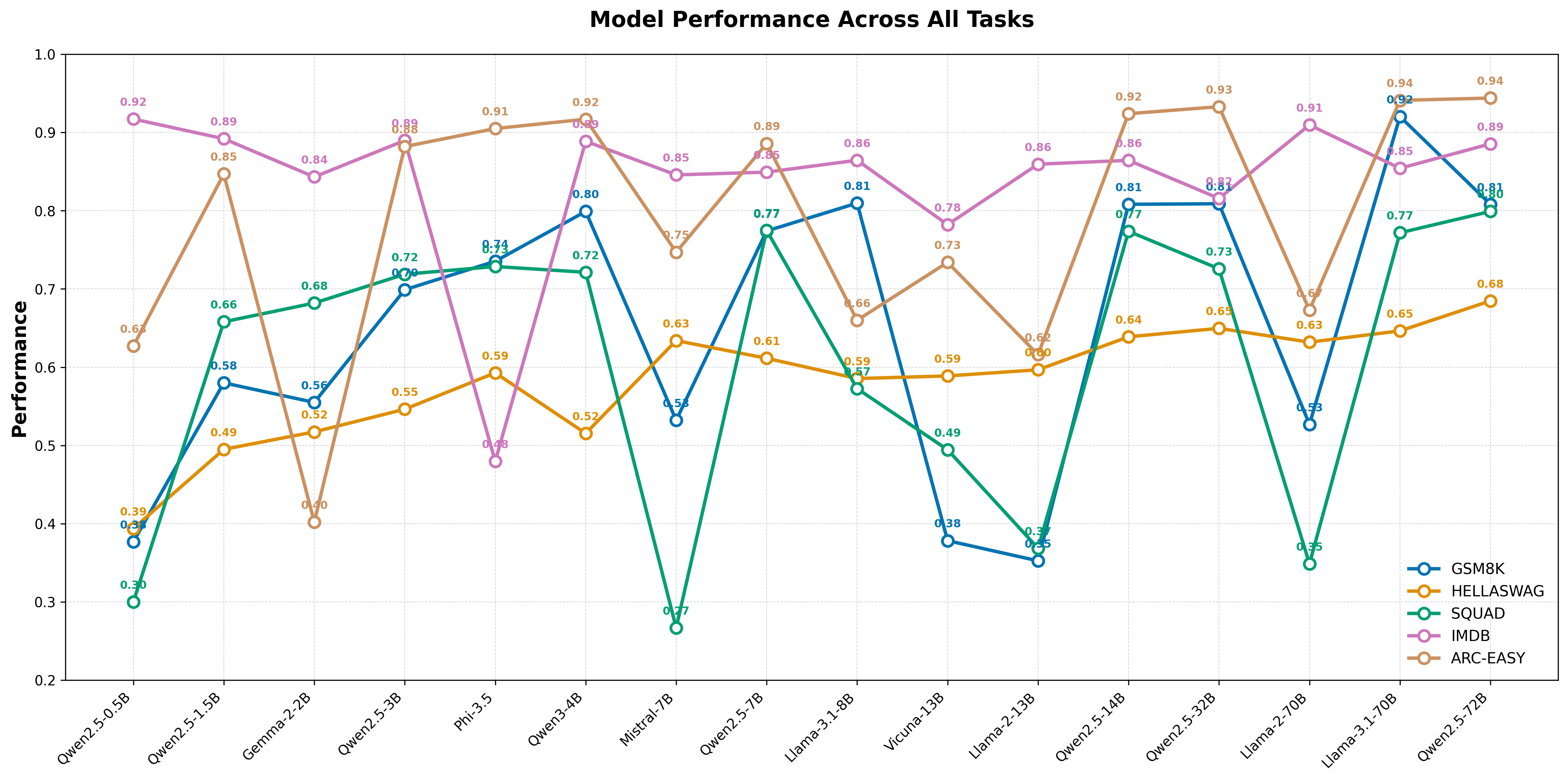}
\caption{Model Accuracy Across All Tasks}
\label{fig:all_tasks_combined}
\end{figure}
All experiments were conducted on NVIDIA A10 and A100 GPUs. To accommodate varying model sizes, we employed multi-GPU tensor parallelism: 0.5-8B models on 1-2 GPUs, 13-15B models on 2-4 GPUs, and 70-72B models on 8 GPUs.

\begin{table}[!ht]
\centering
\small  
\renewcommand{\arraystretch}{0.85}  
\setlength{\tabcolsep}{3pt}
\resizebox{\textwidth}{!}{%
\begin{tabular}{@{}l|ccccc|ccccc@{}}
\hline
\multirow{2}{*}{\textbf{Model}} & \multicolumn{5}{c|}{\textbf{GSM8K}} & \multicolumn{5}{c@{}}{\textbf{SQuAD 2.0}} \\
\cline{2-11}
& \textbf{Acc} & \shortstack{\textbf{Thr}\\(tokens/s)} & \shortstack{\textbf{Lat}\\(ms/token)} & \textbf{Mem(MB)} & \textbf{PER} & \textbf{Acc} & \shortstack{\textbf{Thr}\\(tokens/s)} & \shortstack{\textbf{Lat}\\(ms/token)} & \textbf{Mem(MB)} & \textbf{PER} \\
\hline
Qwen2.5-0.5B & 0.3768 & \textbf{7927} & \textbf{0.15} & \textbf{1024} & 0.4546 & 0.30 & \textbf{854} & \textbf{1.17} & \textbf{1024} & 0.50 \\
Qwen2.5-1.5B & 0.58 & 4417 & 0.25 & 3072 & \textbf{0.6617} & 0.66 & 353 & 2.83 & 3072 & 0.67 \\
Gemma-2-2B & 0.555 & 1761 & 0.66 & 4096 & 0.4887 & 0.68 & 433 & 2.31 & 4096 & \textbf{0.72} \\
Qwen2.5-3B & 0.699 & 2657 & 0.42 & 6144 & 0.6056 & 0.72 & 219 & 4.58 & 6144 & 0.53 \\
Phi-3.5 & 0.7354 & 979 & 1.16 & 7782 & 0.4427 & 0.73 & 172 & 5.83 & 7782 & 0.45 \\
Qwen3-4B & 0.7991 & 1738 & 0.69 & 8192 & 0.5447 & 0.72 & 155 & 6.45 & 8192 & 0.41 \\
Mistral-7B & 0.5322 & 3907 & 1.53 & 7168 & 0.2837 & 0.27 & 127 & 3.93 & 7168 & 0 \\
Qwen2.5-7B & 0.7741 & 374 & 1.46 & 7168 & 0.3491 & 0.77 & 82 & 6.02 & 7168 & 0.37 \\
Llama-3.1-8B & 0.8097 & 151 & 3.74 & 8192 & 0.1957 & 0.57 & 89 & 5.63 & 8192 & 0.33 \\
Llama-2-13B & 0.3525 & 115 & 2.38 & 6656 & 0 & 0.37 & 73. & 3.43 & 6656 & 0.28 \\
Vicuna-13B & 0.3783 & 77 & 3.61 & 6656 & 0.0811 & 0.49 & 53 & 2.37 & 3328 & 0.33 \\
Qwen2.5-14B & 0.8082 & 179 & 1.53 & 7168 & 0.2877 & 0.77 & 28 & 9.08 & 7168 & 0 \\
Qwen2.5-32B & 0.8089 & 293 & 0.47 & 8192 & 0.3482 & 0.73 & 37 & 3.36 & 8192 & 0 \\
Llama-2-70B & 0.5269 & 47 & 2.9 & 17920 & 0.053 & 0.35 & 29 & 4.39 & 17920 & 0.07 \\
Llama-3.1-70B & \textbf{0.9204} & 48 & 2.79 & 17920 & 0.0738 & 0.77 & 19 & 6.31 & 17920 & 0.03 \\
Qwen2.5-72B & 0.8082 & 29 & 4.65 & 18432 & 0 & \textbf{0.80} & 19 & 6.33 & 18432 & 0 \\
\hline
\end{tabular}%
}
\caption{Generation Tasks PER Results Sorted by Model Size}
\label{tab:gsm8k_squad_results}
\end{table}

\begin{table}[!ht]
\centering
\resizebox{\textwidth}{!}{%
\begin{tabular}{l|ccccc|ccccc|ccccc}
\hline
\multirow{2}{*}{\textbf{Model}} & \multicolumn{5}{c|}{\textbf{HellaSwag}} & \multicolumn{5}{c|}{\textbf{IMDB}} & \multicolumn{5}{c}{\textbf{ARC-Easy}} \\
\cline{2-16}
& \textbf{Acc} & \shortstack{\textbf{Thr}\\(samples/s)} & \shortstack{\textbf{Lat}\\(ms/sample)} & \textbf{Mem(MB)} & \textbf{PER} & \textbf{Acc} & \shortstack{\textbf{Thr}\\(samples/s)} & \shortstack{\textbf{Lat}\\(ms/sample)} & \textbf{Mem(MB)} & \textbf{PER} & \textbf{Acc} & \shortstack{\textbf{Thr}\\(samples/s)} & \shortstack{\textbf{Lat}\\(ms/sample)} & \textbf{Mem(MB)} & \textbf{PER} \\
\hline
Qwen2.5-0.5B & 0.39 & \textbf{65} & \textbf{15} & \textbf{1024} & 0 & \textbf{0.92} & \textbf{193} & \textbf{5} & \textbf{1024} & \textbf{1.00} & 0.63 & \textbf{574} & \textbf{2} & \textbf{1024} & \textbf{0.80} \\
Qwen2.5-1.5B & 0.50 & 36 & 28 & 3072 & \textbf{0.63} & 0.89 & 66 & 15 & 3072 & 0.71 & 0.85 & 178 & 6 & 3072 & 0.68 \\
Gemma-2-2B & 0.52 & 21 & 48 & 4096 & 0.53 & 0.84 & 20 & 50 & 4096 & 0.46 & 0.40 & 87 & 11 & 4096 & 0 \\
Qwen2.5-3B & 0.55 & 22 & 45 & 6144 & 0.57 & 0.89 & 22 & 45 & 6144 & 0.48 & 0.88 & 60 & 17 & 6144 & 0.47 \\
Phi-3.5 & 0.59 & 20 & 51 & 7782 & 0.56 & 0.48 & 6 & 154 & 7782 & 0 & 0.91 & 42 & 24 & 7782 & 0.40 \\
Qwen3-4B & 0.52 & 18 & 56 & 8192 & 0.48 & 0.89 & 19 & 53 & 8192 & 0.42 & 0.92 & 47 & 21 & 8192 & 0.42 \\
Qwen2.5-7B & 0.61 & 11 & 87 & 14336 & 0.37 & 0.85 & 3 & 153 & 7168 & 0.07 & 0.89 & 15 & 33 & 7168 & 0.28 \\
Mistral-7B & 0.61 & 5 & 104 & 7168 & 0.37 & 0.85 & 11 & 93 & 14336 & 0.25 & 0.75 & 14 & 35 & 7168 & 0.25 \\
Llama-3.1-8B & 0.59 & 4 & 132 & 8192 & 0.30 & 0.86 & 5 & 96 & 8192 & 0.24 & 0.66 & 7 & 38 & 4096 & 0.17 \\
Vicuna-13B & 0.59 & 10 & 25 & 6656 & 0.49 & 0.78 & 17 & 15 & 6656 & 0.43 & 0.73 & 4 & 66 & 6656 & 0.04 \\
Llama-2-13B & 0.60 & 9 & 27 & 6656 & 0.49 & 0.86 & 14 & 18 & 6656 & 0.43 & 0.62 & 5 & 53 & 6656 & 0.10 \\
Qwen2.5-14B & 0.64 & 1 & 234 & 7168 & 0 & 0.86 & 10 & 24 & 7168 & 0.38 & 0.92 & 4 & 68 & 7168 & 0 \\
Qwen2.5-32B & 0.65 & 5 & 26 & 8192 & 0.41 & 0.82 & 5 & 46 & 16384 & 0.18 & 0.93 & 11 & 12 & 8192 & 0.28 \\
Llama-2-70B & 0.66 & 2 & 53 & 17920 & 0.14 & 0.91 & 3 & 45 & 17920 & 0.09 & 0.67 & 5 & 23 & 17920 & 0.07 \\
Llama-3.1-70B & 0.65 & 3 & 49 & 17920 & 0.15 & 0.85 & 3 & 39 & 17920 & 0.10 & 0.94 & 6 & 20 & 17920 & 0.10 \\
Qwen2.5-72B & \textbf{0.68} & 3 & 44 & 18432 & 0 & 0.89 & 2 & 57 & 18432 & 0 & \textbf{0.94} & 6 & 21 & 18432 & 0 \\
\hline
\end{tabular}%
}

\caption{Classification Task PER Results Sorted by Model Size}
\label{tab:hellaswag_imdb_arceasy_results}
\end{table}

\subsection{Task-Specific Scaling Characteristics}

As illustrated in Figure \ref{fig:all_tasks_combined}, we observe three distinct scaling regimes across benchmark tasks. Simple classification tasks (IMDB) exhibit saturation at sub-billion scales: Qwen2.5-0.5B achieves 91.7\% accuracy with minimal gains from larger models (Qwen2.5-72B: 88.6\%). Mathematical reasoning (GSM8K) demonstrates pronounced but non-monotonic scaling benefits: accuracy improves from 37.7\% (0.5B) to 92.0\% (Llama-3.1-70B), yet Qwen2.5-14B (80.8\%) achieves 87.8\% of peak performance using only 20\% of parameters. Scientific reasoning (ARC-Easy) and commonsense understanding (HellaSwag) show substantial gains from 0.5B to 3--7B (HellaSwag: 39.4\% $\to$ 63.4\%, ARC-Easy: 62.7\% $\to$ 88.2\%) with diminishing returns thereafter.

\subsection{Generation Task Efficiency}

Table \ref{tab:gsm8k_squad_results} reveals small models (1.5--3B) consistently achieve the highest PER scores (0.60--0.72) for generation tasks, dramatically outperforming larger alternatives. Despite 10--35\% accuracy trade-offs versus 70B+ models, compact models deliver 40--90$\times$ superior per-GPU throughput (350--4400 tokens/s/GPU vs. 20--50 tokens/s/GPU). Models exceeding 14B parameters exhibit PER scores below 0.35, with 70B+ models falling below 0.08. Each parameter doubling beyond 3B yields diminishing accuracy returns (5--15\% improvement) while incurring superlinear efficiency penalties.

\subsection{Classification Task Efficiency}

Table \ref{tab:hellaswag_imdb_arceasy_results} demonstrates ultra-small models (0.5--2B) dominate classification PER rankings. Small models process data faster per GPU while maintaining competitive accuracy. Task-dependent characteristics emerge: sentiment classification (IMDB) saturates at 0.5B (91.7\%) with negligible gains from scaling, while scientific reasoning (ARC-Easy) shows improvements from 0.5B (62.7\%) to 72B (94.4\%). However, even for complex classification tasks, small models' efficiency gains (50\% improvement) substantially exceed accuracy trade-offs, establishing them as optimal for cost-efficient pipelines.

\section{Conclusion}
Small models (0.5--3B) achieve superior Performance-Efficiency Ratios (PER) over larger counterparts, which exhibit diminishing returns beyond 14B parameters. While medium-scale models (3--14B) offer an optimal performance balance, 0.5--3B models are most effective for resource-constrained deployments. These findings provide a quantitative basis for task-specific selection, challenging the ``bigger is better'' paradigm.

\begin{footnotesize}

\bibliographystyle{unsrt}


\ifarxiv
\bibliography{citations,citations_arxiv}
\else
\bibliography{citations}
\fi

\end{footnotesize}


\end{document}